\documentclass[11pt,a4paper]{article}

\usepackage{etex}
\usepackage[utf8]{inputenc}
\usepackage{amssymb, amsmath, amsfonts} 
\usepackage{graphicx}
\usepackage{latexsym}
\usepackage{multirow}
\usepackage{empheq,fancybox}
\usepackage{subfig}
\usepackage[figuresright]{rotating}
\usepackage{tabularx}
\usepackage[table]{xcolor}
\usepackage{pstricks,pstricks-add,pst-node,pst-plot,pst-coil}
\usepackage{algorithm}
\usepackage{algorithmic}
\usepackage[algo2e]{algorithm2e} 
\usepackage{tabularx}
\usepackage{booktabs}
\usepackage{url}
\usepackage{csquotes}
\usepackage{rotating}
\usepackage{bigstrut}

\usepackage[draft,nomargin,inline]{fixme}
\fxsetface{inline}{\itshape}
\fxsetface{env}{\itshape}
\fxusetheme{color}

\newcommand{\drate}{d_{\mathrm{rate}}}

\newcommand{\NULL}{\textsc{null}}

\newcommand{\numsol}{\ensuremath{n_{\mathrm{a}}}}
\newcommand{\gencmsabip}{\textsc{Gen-Cmsa-Bip}}
\newcommand{\gencpcmsabip}{\textsc{Gen/Cp-Cmsa-Bip}}
\newcommand{\cplexopt}{\textsc{Cplex-Opt}}
\newcommand{\cplexheur}{\textsc{Cplex-Heur}}

\begin{document}

\title{Generic CP-Supported CMSA for Binary Integer Linear Programs}

\author{Christian Blum$^{1}$ and Haroldo Gambini Santos$^{2}$ \\
~\\
$^1$Artificial Intelligence Research Institute (IIIA-CSIC)\\
Campus of the UAB, Bellaterra, Spain\\
{\sf christian.blum@iiia.csic.es}\\
~\\
$^2$Department of Computer Science\\
Universidade Federal de Ouro Preto, Brazil\\
{\sf haroldo@ufop.edu.br}}

\date{}

\maketitle

\begin{abstract}
Construct, Merge, Solve \& Adapt (CMSA) is a general hybrid metaheuristic for solving combinatorial optimization problems. At each iteration, CMSA (1) constructs feasible solutions to the tackled problem instance in a probabilistic way and (2) solves a reduced problem instance (if possible) to optimality. The construction of feasible solutions is hereby problem-specific, usually involving a fast greedy heuristic. The goal of this paper is to design a problem-agnostic CMSA variant whose exclusive input is an integer linear program (ILP). In order to reduce the complexity of this task, the current study is restricted to binary ILPs. In addition to a basic problem-agnostic CMSA variant, we also present an extended version that makes use of a constraint propagation engine for constructing solutions. The results show that our technique is able to match the upper bounds of the standalone application of CPLEX in the context of rather easy-to-solve instances, while it generally outperforms the standalone application of CPLEX in the context of hard instances. Moreover, the results indicate that the support of the constraint propagation engine is useful in the context of problems for which finding feasible solutions is rather difficult.
\end{abstract}
\section{Introduction}

Construct, Merge, Solve \& Adapt (CMSA)~\cite{blum2016construct} is a hybrid metaheuristic that can be applied to any combinatorial optimization problem for which is known a way of generating feasible solutions, and whose subproblems can be solved to optimality by a black-box solver. Moreover, note that CMSA is thought for those problem instances for which the application of the standalone black-box solver is not feasible due to the problem instance size and/or difficulty. The main idea of CMSA is to generate reduced sub-instances of the original problem instances, based on feasible solutions that are constructed at each iteration, and to solve these reduced instances by means of the black-box solver. Obviously, the parameters of CMSA have to be adjusted in order for the size of the reduced sub-instances to be such that the black-box solver can solve them efficiently. CMSA has been applied to several NP-hard combinatorial optimization problems, including minimum common string partition~\cite{blum2016construct,Blum2016:hm}, the repetition-free longest common subsequence problem~\cite{Blum2017:rflcs-joh}, and the multi-dimensional knapsack problem~\cite{LBB2017:evocop}. \\

A possible disadvantage of CMSA is the fact that a problem-specific way of probabilistically generating solutions is used in the above-mentioned applications. Therefore, the goal of this paper is to design a CMSA variant that can be easily applied to different combinatorial optimization problems. One way of achieving this goal is the development of a solver for a quite general problem. Combinatorial optimization problems can be conveniently expressed as Integer Linear Programs (ILPs) in the format $\textrm{min } \vec{c}^T\vec{x} : A\vec{x}=\vec{b}, \vec{x} \in \mathbb{Z}^n$, where $A$ indicates a constraints matrix, $\vec{b}$ and $\vec{c}$ are the cost and right-hand-side vectors, respectively and $\vec{x}$ is a vector of decision variables whose values are restricted to integral numbers. In this paper we propose a generic CMSA for binary integer programs (BIPs), that are obtained when $\vec{x}\in \{0,1\}^n$. This type of problem is generic enough to model a wide range of combinatorial optimization problems, from the classical traveling salesman problem~\cite{applegate2006traveling} to protein threading problems~\cite{Xy2003} and a myriad of applications listed in the MIPLIB 2010 collection of problem instances~\cite{koch2011miplib}. As CMSA is an algorithm that makes use of a solution construction mechanism at each iteration, one of the challenges that we address in this paper is the fast production of feasible solutions for general BIPs. For this purpose we support the proposed generic CMSA with a constraint propagation (CP) tool for increasing the probability to generate feasible solutions.

This paper is organized as follows. The next section discusses related work. In Section~\ref{sec:cmsa}, the original version of CMSA is presented, which assumes that the type of the tackled problem is known. The generic CMSA proposal for general BIPs is described in Section~\ref{sec:algo}. Finally, an extensive experimental evaluation is presented in Section~\ref{sec:comparison} and conclusions as well as an outlook to future work are provided in Section~\ref{sec:conclusions}.

\section{Related Works}

The development of fast, reliable general purpose combinatorial optimization solvers is a topic that occupies operations research practitioners since many years. The main reason being that the structure of real world optimization problems usually does not remain fixed over time: constraints usually change over time and solvers optimized for a very particular problem structure may lose their efficiency in this process. Thus, a remarkable interest in integer linear programming (ILP) software packages exists, with several commercially successful products such as IBM CPLEX, Gurobi and XPRESS. This success can be attributed to the continuous improvements concerning the performance of these solvers~\cite{Johnson2000,Gamrath2013} and the availability of high level languages such AMPL~\cite{fourer1993ampl}. The application of these solvers to instances with very different structures creates many challenges. From a practical point of view, the most important one is,  possibly, the ability of quickly providing high quality feasible solutions: even though a complete search is executed, it is quite often the case that time limits need to be imposed and a truncated search is performed. Thus, several methods have been proposed to try to produce feasible solutions in the first steps of the search process. One of the best known approaches is the so-called \emph{feasibility pump}~\cite{fischetti2005feasibility,Fischetti2009}.

In the context of metaheuristics, Kochenberger et al.~\cite{kochenberger2014unconstrained} developed a general solver for unconstrained binary quadratic programming (UBQP) problems. A whole range of important combinatorial optimization problems such as set partitioning and $k$-coloring can be easily modeled as special cases of the UBQP problem. Experiments showed that their general solver was able to produce high quality solutions much faster than the general purpose ILP solver CPLEX for hard problems such as the the set partitioning problem. Brito \& Santos~\cite{SGM2014:hm} proposed a local search approach for solving BIPs, obtaining some encouraging results when comparing to the COIN-OR CBC Branch-and-Cut solver. In the context of constraint programming, Benoist et al.~\cite{benoist2011localsolver} proposed a fast heuristic solver (LocalSolver) based on local search. Experiments showed that LocalSolver outperformed several other solvers, especially for what concerns executions with very restricted computation times. In this paper we propose a CMSA solver for solving BIPs. This format is more restricted than the LocalSolver format, where non-linear functions can be used, but  much more general than the UBQP, which can be easily modeled as a special case of binary programming. One advantage of BIPs is that several high performance solvers can be used to solve the sub-problems generated within CMSA, a feature that will be explored in the next sections.

\section{Original CMSA in the Context of BIPs}
\label{sec:cmsa}

As already mentioned, in this work we focus on solving BIPs. Any BIP can be expressed in the following way:
\begin{equation}\label{eqn:bip}
   \min \{\vec{c}^T\vec{x} : A\vec{x} \leq \vec{b}, x_j \in \{0, 1\} \; \forall j = 1, \ldots, n\}
\end{equation}
where $A$ is an $m \times n$ matrix, $\vec{b}$ is the right-hand-size vector of size $m$, $\vec{c}$ is a cost vector, and $\vec{x}$ is the vector of $n$ binary decision variables. Note that $m$ is the number of constraints of this BIP. 

In the following we describe the original CMSA algorithm from~\cite{blum2016construct}. However, instead of providing a general description as in~\cite{blum2016construct}, our description is already tailored for the application to BIPs. In order to clarify this fact, the algorithm described below is labeled CMSA-BIP. In general, the main idea of CMSA algorithms is to take profit from an efficient complete solver even in the context of problem instances that are too large to be solved directly. The general idea of CMSA is as follows. At each iteration, CMSA probabilistically generates solutions to the tackled problem instance. Next, the solution components that are found in these solutions are added to a sub-instance of the original problem instance. Subsequently, an exact solver is used to solve the sub-instance (if possible in the given time) to optimality.\footnote{In the context of problems that can be modelled as BIPs, any black-box ILP solver such as, for example, CPLEX can be used for this purpose} Moreover, the algorithm is equipped with a mechanism for deleting seemingly useless solution components from the sub-instance. This is done such that the sub-instance has a moderate size and can be solved rather quickly to optimality. \\

In the context of CMSA-BIP, any combination of a variable $x_j$ with one of its values $v \in \{0, 1\}$ is a solution component denoted by $(x_j, v)$. Given a BIP instance, the complete set of solution components if denoted by $C$. Any sub-instance of the given BIP is a subset $C'$ of $C$, that is, $C' \subseteq C$. Such a sub-instance $C' \subseteq C$ is feasible, if $C'$ contains for each variable $x_j$ ($j = 1, \ldots, n$) at least one solution component $(x_j, v)$, that is, either $(x_j, 0)$, or $(x_j, 1)$, or both. Moreover, a solution to the given BIP is any binary vector $\vec{s}$ that fulfills the constraints from Eq.~(\ref{eqn:bip}). Note that a feasible solution $\vec{s}$ contains $n$ solution components: $\{(x_j, s_j) \mid j=1,\ldots,n\}$. 

The pseudo-code of the CMSA-BIP algorithm is given in Algorithm~\ref{algo:cmsa}. At each iteration the following is done. First, the best-so-far solution $\vec{s}^{\mathrm{bsf}}$ is initialized to $\NULL$, indicating that no such solution exists yet. Moreover, sub-instance $C'$ is initialized to the empty set. Note, also, that each solution component $(x_j, v) \in C$ has a so-called age value denoted by $\mathit{age}[(x_j, v)]$. All these age values are initialized to zero at the start of the algorithm. Then, at each iteration, $\numsol$ solutions are probabilistically generated in function \textsf{ProbabilisticSolutionGeneration}($C$); see line 6 of Algorithm~\ref{algo:cmsa}. As mentioned above, problem-specific heuristics are generally used for this purpose. The solution components found in the constructed solutions are then added to $C'$. Next, an ILP solver is applied in function \textsf{ApplyILPSolver}($C'$) to find a possibly optimal solution $\vec{s}^{\prime}_{\mathrm{opt}}$ to the restricted problem instance $C'$ (see below for a more detailed description). Note that $\NULL$ is returned in case the ILP solver cannot find any feasible solution. If $\vec{s}^{\prime}_{\mathrm{opt}}$ is better than the current best-so-far solution $\vec{s}^{\mathrm{bsf}}$, solution $\vec{s}^{\prime}_{\mathrm{opt}}$ is taken as the new best-so-far solution. Next, sub-instance $C'$ is adapted on the basis of solution $\vec{s}^{\prime}_{\mathrm{opt}}$ in conjunction with the age values of the solution components; see function \textsf{Adapt}($C'$, $\vec{s}^{\prime}_{\mathrm{opt}}$, $\mathit{age}_{\max}$) in line 14. This is done as follows. First, the age of all solution components from $C'$ that are not in $\vec{s}^{\prime}_{\mathrm{opt}}$ is incremented. Moreover, the age of each solution component from $\vec{s}^{\prime}_{\mathrm{opt}}$ is re-initialized to zero. Subsequently, those solution components from $C'$ with an age value greater than $\mathit{age}_{\max}$---which is a parameter of the algorithm---are removed from $C'$. This causes that solution components that never appear in solutions derived by the ILP solver do not slow down the solver in subsequent iterations. On the other side, components which appear in the solutions returned by the ILP solver should be maintained in $C'$.

\begin{algorithm}[t]
\caption{CMSA-BIP: CMSA for solving BIPs} \label{algo:cmsa}
\begin{algorithmic}[1]
\STATE {\bf given:} a BIP instance, and values for the algorithm parameters
\STATE $\vec{s}^{\mathrm{bsf}} := \NULL$; $C' := \emptyset$
\STATE $\mathit{age}[(x_j, v)] := 0$ for all $(x_j, v) \in C$
\WHILE{CPU time limit not reached}
    \FOR{$i = 1, \ldots, \numsol$}
        \STATE $\vec{s} :=$ \textsf{ProbabilisticSolutionGeneration}($C$)
        \FOR{$j=1,\ldots,n$}
            \IF{$(x_j, s_j) \notin C'$}
                \STATE $\mathit{age}[(x_j, s_j)] := 0$
                \STATE $C' := C' \cup \{(x_j, s_j)\}$
            \ENDIF
        \ENDFOR
    \ENDFOR
    \STATE $\vec{s}^{\prime}_{\mathrm{opt}} :=$ \textsf{ApplyILPSolver}($C'$)
    \STATE {\bf if} $\vec{s}^{\prime}_{\mathrm{opt}}$ is better than $\vec{s}^{\mathrm{bsf}}$ {\bf then} $\vec{s}^{\mathrm{bsf}} := \vec{s}^{\prime}_{\mathrm{opt}}$ {\bf end if}
    \STATE \textsf{Adapt}($C'$, $\vec{s}^{\prime}_{\mathrm{opt}}$, $\mathit{age}_{\max}$)
\ENDWHILE
\RETURN $\vec{s}^{\mathrm{bsf}}$
\end{algorithmic}
\end{algorithm}

Finally, the BIP that is solved at each iteration in function \textsf{ApplyILPSolver}($C'$) is generated by adding the following constraints to the original BIP. For each $j = 1,\ldots, n$ the following is done. If $C'$ only contains solution component $(x_j, 0)$, the additional constraint $x_j = 0$ is added to the original BIP. Otherwise, if $C'$ only contains solution component $(x_j, 1)$, the additional constraint $x_j = 1$ is added to the original BIP. Nothing is added to the original BIP in case $C'$ contains both solution components. Note that the ILP solver is applied with a computation time limit of $t^{SUB}$ CPU seconds, which is a parameter of the algorithm.

\section{Generic Way of Generating Solutions for BIPs}
\label{sec:algo}

In those cases in which the optimization problem modeled by the given BIP is not known, we need a generic way of generating solutions to the given BIP in order to be able to apply the CMSA-BIP algorithm described in the previous section. In the following we first describe a basic solution construction mechanims, and afterwards an alternative mechanism which uses a CP tool for increasing the probability to generate feasible solutions. The first algorithm variant is henceforth denoted as \textsc{Gen-Cmsa-Bip} (standing for generic CMSA-BIP) and the second algorithm variant as \textsc{Gen/Cp-Cmsa-Bip} (standing for generic CMSA-BIP with CP support).

\subsection{Basic Solution Construction Mechanism}
\label{sec:sol-constr}

Before starting with the first CMSA-BIP iteration, a node heuristic of the applied ILP solver might be used in order to obtain a first feasible solution. In our case, we used the node heuristic of CPLEX. If, in this way, a feasible solution can be obtained it is stored in $\vec{s}^{\mathrm{bsf}}$. Otherwise, $\vec{s}^{\mathrm{bsf}}$ is set to $\NULL$. If, after this step, $\vec{s}^{\mathrm{bsf}}$ has value $\NULL$, the LP relaxation of the given BIP is solved. However, in order not to spend too much computation time on this step, a computation time limit of $t^{LP}$ seconds is applied. After this, the possibly optimal solution of the LP relaxation is stored in vector $\vec{x}^{LP}$. Then, whenever function \textsf{ProbabilisticSolutionGeneration}($C$) is called, the following is done. First, a so-called sampling vector $\vec{x}^{\mathrm{samp}}$ for sampling new (possibly feasible) solutions by randomized rounding is generated. If $\vec{s}^{\mathrm{bsf}} \not= \NULL$, $\vec{x}^{\mathrm{samp}}$ is generated based on $\vec{s}^{\mathrm{bsf}}$ and a so-called \emph{determinism rate} $0 < d_{\mathrm{rate}} < 0.5$ as follows:
\[ x^{\mathrm{samp}}_j =
  \begin{cases}
    \drate                           & \quad \text{if } s^{\mathrm{bsf}}_j = 0 \\
    1-\drate                         & \quad \text{if } s^{\mathrm{bsf}}_j = 1
  \end{cases}
\]
for all $j=1,\ldots,n$. In case $\vec{s}^{\mathrm{bsf}} = \NULL$, $\vec{x}_{\mathrm{samp}}$ is---for all $j=1,\ldots,n$---generated on the basis of $\vec{x}^{LP}$:
\[ x^{\mathrm{samp}}_j =
  \begin{cases}
    x^{LP}_j       & \quad \text{if } \drate \leq x^{LP}_j \leq 1-\drate \\
    \drate         & \quad \text{if } x^{LP}_j < \drate \\
    1-\drate       & \quad \text{if } x^{LP}_j > 1 - \drate
  \end{cases}
\]
After generating $\vec{x}_{\mathrm{samp}}$, a possibly infeasible binary solutions $\vec{s}$ is generated from $\vec{x}_{\mathrm{samp}}$ by randomized rounding. Note that this is done in the order $j=1,\ldots,n$.

\subsection{CP Supported Construction Mechanism}

Our algorithm makes use of the Constraint Propagation (CP) engine \texttt{cprop} that implements ideas from~\cite{Achterberg2007,Sandholm2006}.\footnote{The used CP tool can be obtained at \url{https://github.com/h-g-s/cprop}.} The support of CP is used in the following two ways. First, all constraints are processed and implications derived from the constraint set are detected and the problem is preprocessed to keep those variables fixed throughout the search process. Second, the solution construction mechanism changes in the following way. Instead of deriving values for the variables in the order $j=1,\ldots,n$, a random order $\pi$ is chosen for each solution construction. That is, at step $j$, instead of deriving a value for variable $x_j$, instead a value for variable $x_{\pi(j)}$ is derived. Then, after deciding for a value for variable $x_{\pi(j)}$, the CP tool checks if this assignment will produce an infeasible solution. If this is the case, variable $x_{\pi(j)}$ is fixed to the alternative value. If, again, the CP tool determines that this setting cannot lead to a feasible solution, the solution construction proceeds as described in Section~\ref{sec:sol-constr}, that is, finalizing the solution construction without further CP support. Otherwise---that is, if a feasible value can be chosen for the current variable---the CP might indicate possible implications consisting of further variables that, as a consequence, have to be fixed to certain values. All these implications are dealt with, before dealing with the next non-fixed variable according to $\pi$.\footnote{Note that, after fixing a value for $x_{\pi(j)}$, the value of $x_{\pi(j+1)}$ might already be fixed due to one of the implications dealt with earlier.}
%
%

\subsection{An Additional Algorithmic Aspect}

Instead of using fixed values for CMSA-BIP parameters $\drate$ and $t^{SUB}$, we implemented the following scheme. For both parameters we use a lower bound and an upper bound. At the start of CMSA-BIP, the values of $\drate$ and $t^{SUB}$ are set to the lower bound. Whenever an iteration improves $\vec{s}^{\mathrm{bsf}}$, the values of $\drate$ and $t_{SUB}$ are set back to their respective lower bounds. Otherwise, the values of $\drate$ and $t_{SUB}$ are increased by a factor determined by substracting the lower bound value from the upper bound value and dividing the result by 5.0. Finally, whenever the value of $\drate$, respectively $t_{SUB}$, exceeds its upper bound, it is set back to the lower bound value. This procedure is inspired by variable neighborhood search (VNS)~\cite{mladenovic1997variable}.

\section{Experimental Evaluation}
\label{sec:comparison}

In the following we present an experimental evaluation of \gencmsabip\ and \gencpcmsabip\ in comparison to the standalone application of the ILP solver IBM ILOG CPLEX v12.7. Note that the same version of CPLEX was applied within both CMSA variants. Moreover, in all cases CPLEX was executed in one-threaded mode. In order to ensure a fair comparison, CPLEX was executed with two different parameter settings in the standalone mode: the default parameter settings, and with the MIP emphasis parameter set to a value of 4 (which means that the focus of CPLEX is on finding good solutions rather than on proving optimality). The default version of CPLEX is henceforth denoted by \cplexopt\ and the heuristic version of CPLEX by \cplexheur. All techniques were implemented in ANSI C++ (with the Concert Library of ILOG for implementing everything related to the ILP models), and using GCC 5.4.0 for compiling the software. Moreover, the experimental evaluation was performed on a cluster of PCs with Intel(R) Xeon(R) CPU 5670 CPUs of 12 nuclei of 2933 MHz and at least 40 Gigabytes of RAM. 

\subsection{Considered Problem Instances}

\begin{table}[!t]
\caption{Characteristics of the 30 BIPs that were considered.}
\label{tab:instance-characteristics}
\centering
\scalebox{0.9}{
\begin{tabular}{lrrrr} \hline
{\bf BIP instance name}      &   {\bf \# Cols/Vars} & {\bf \# Rows} & {\bf Opt.~Val.} & {\bf MIPLIB status} \\ \hline
\texttt{acc-tight5}          &       1339 &    3052 & 0.0                 & Easy \\
\texttt{air04}               &       8904 &     823 & 56137.0             & Easy \\
\texttt{cov1075}             &        120 &     637 & 20.0                & Easy \\
\texttt{eilB101}             &       2818 &     100 & 1216.92             & Easy \\
\texttt{ex9}                 &      10404 &   40962 & 81.0                & Easy \\
\texttt{netdiversion}        &     129180 &  119589 & 242.0               & Easy \\ 
\texttt{opm2-z7-s2}          &       2023 &   31798 & -10280.0            & Easy \\
\texttt{tanglegram1}         &      34759 &   68342 & 5182.0              & Easy \\
\texttt{vpphard}             &      51471 &   47280 & 5.0                 & Easy \\ \hline
\texttt{ivu52}               &     157591 &    2116 & 481.007             & Hard \\
\texttt{opm2-z12-s14}        &      10800 &  319508 & -64291.0            & Hard \\
\texttt{p6b}                 &        462 &    5852 & -63.0               & Hard \\
\texttt{protfold}            &       1835 &    2112 & -31.0               & Hard \\
\texttt{queens-30}           &        900 &     960 & -40.0               & Hard \\
\texttt{reblock354}          &       3540 &   19906 & -39280521.23        & Hard \\
\texttt{rmine10}             &       8439 &   65274 & -1913.88            & Hard \\
\texttt{seymour-disj-10}     &       1209 &    5108 & 287.0               & Hard \\
\texttt{wnq-n100-mw99-14}    &      10000 &  656900 & 259.0               & Hard \\ \hline
\texttt{bab1}                &      61152 &   60680 & Unknown             & Open \\
\texttt{methanosarcina}      &       7930 &   14604 & Unknown             & Open \\
\texttt{ramos3}              &       2187 &    2187 & Unknown             & Open \\
\texttt{rmine14}             &      32205 &  268535 & Unknown             & Open \\
\texttt{rmine25}             &     326599 & 2953849 & Unknown             & Open \\
\texttt{sts405}              &        405 &   27270 & Unknown             & Open \\ 	
\texttt{sts729}              &        729 &   88452 & Unknown             & Open \\
\texttt{t1717}               &      73885 &     551 & Unknown             & Open \\
\texttt{t1722}               &      36630 &     338 & Unknown             & Open \\ \hline
\texttt{mcsp-2000-4}         &    1335342 &    4000 & Unkonwn             & n.a. \\
\texttt{rflcs-2048-3n-div-8} &       5461 & 7480548 & Unknown             & n.a. \\ 
\texttt{rcjs-20testS6}       &     273372 &   29032 & Unknown             & n.a. \\ \hline
\end{tabular}}
\end{table}

The properties of the 30 selected BIPs are described in Table~\ref{tab:instance-characteristics}. The first 27 instances are taken from MIPLIB 2010 (\url{http://miplib.zib.de/miplib2010.php}), which is one of the best-known libraries for integer linear programming. More specifically, the ILPs on MIPLIB are classified into three hardness categories: \emph{easy}, \emph{hard}, and \emph{open}. From each one of the these categories we chose (more or less randomly) 9 BIPs. In addition, we selected three instances from recent applications found in the literature:
\begin{itemize}
  \item \texttt{mcsp-2000-4} is an instance of the minimum common string partition problem with input strings of length 2000 and an alphabet size of four~\cite{blum2016construct}. The hardness of this instance is due to a massive amount of constraints.
  \item \texttt{rflcs-2048-3n-div-8} is an instance of the repetition-free longest common subsequence problem with two input strings of length 2048 and an alphabet size of 768~\cite{Blum2017:rflcs-joh}. This instance is hard to solve due to the large number of variables.
  \item \texttt{rcjs-20testS6} is an instance of the resource constraint job scheduling problem considered in~\cite{6256177}. Finding feasible solutions for this problem is, for general purpose ILP solvers, rather time consuming.
\end{itemize}

\subsection{Parameter Setting}

Both generic CMSA variants have the following parameters for which well-working values must be found: (1) the number of solution constructions per iteration ($\numsol$), (2) the maximumm age of solution components ($\mathit{age}_{\max}$), (3) a computation time limit for solving the LP relaxation ($t^{LP}$), (4) a lower and an upper bound for the determinism rate ($\drate$ (LB) and $\drate$ (UB)), and (5) a lower and an upper bound for the computation time limit of the ILP solver at each iteration ($t^{SUB}$ (LB) and $t^{SUB}$ (UB)). 

Concerning $\mathit{age}_{\max}$, it became clear during preliminary experiments that this parameter has not the same importance for \gencmsabip\ and \gencpcmsabip\ as it has for a problem-specific CMSA. In other words, while a setting of $\numsol = 10$ and $\mathit{age}_{\max} = 3$ is essentially different to a setting of $\numsol = 30$ and $\mathit{age}_{\max} = 1$ for a problem-specific CMSA, this is not the case for the generic CMSA variants. This is related to the way of constructing solutions. In a problem-specific CMSA, the greedy heuristic that is used in a probabilistic way biases the search towards a certain area of the search space. This is generally beneficial, but may have the consequence that some solution components that are needed for high-quality solutions have actually a low probability to be included in the constructed solutions. A setting of $\mathit{age}_{\max} > 1$ provides $\mathit{age}_{\max}$ opportunites---that is, applications of the ILP solver---to find high-quality solutions that incorporate such solution components. In contrast, the way of constructing solutions in the generic CMSA variants does not produce this situtaiton. Therefore, we decided for a setting of $\mathit{age}_{\max} = 1$ for all further experiments. Apart from $\mathit{age}_{\max}$, after preliminary experiments we also fixed the following parameter values:
\begin{itemize}
  \item $\numsol = 5$  
  \item $t^{LP} = 10.0$ 
  \item The lower bound of $t^{SUB}$ is set to 30.0 and the upper bound to 100.0
\end{itemize}
The parameter that has the strongest impact on the performance of the generic CMSA variants is $\drate$. We noticed that both generic CMSA variants are quite sensitive to the setting of the lower and the upper bound for this parameter. However, in order to avoid a fine-tuning for each single problem instance, we decided to identify four representative parameter value configurations in order to cover the characteristics of the 30 selected problem instances. Both generic CMSA variants are then applied with all four parameter configurations to all 30 problem instances. For each problem instance we take the result of the respective best configuration as the final result (and we indicate with which configuration this result was obtained). The four utilized parameter configurations are described in Table~\ref{tab:parameter-values}.

\begin{table}[!t]
\caption{The four parameter configurations used for both \gencmsabip and \gencpcmsabip.}
\label{tab:parameter-values}
\centering
\scalebox{1.0}{
\begin{tabular}{lrr} \hline
{\bf Parameter Configuration}      &  $\drate$ (LB) & $\drate$ (UB) \\ \hline
Configuration 1                    &  0.03          & 0.08          \\
Configuration 2                    &  0.05          & 0.15          \\
Configuration 3                    &  0.1           & 0.3           \\
Configuration 4                    &  0.3           & 0.5           \\ \hline
\end{tabular}}
\end{table}

\subsection{Results}

All four approaches (\gencmsabip, \gencpcmsabip, \cplexopt, and \cplexheur) were applied with a computation time limit of 1000 CPU seconds to each one of the 30 problem instances. However, as \gencmsabip\ and \gencpcmsabip\ are stochastic algorithms, they are applied 10 times to each instance, while the two CPLEX variants are applied exactly once to each instance. The numerical results are provided in Table~\ref{tab:results:bips}, which has the following structure. The first column contains the problem instance name, and the second column provides the value of an optimal solution (if known).\footnote{Note that all considered BIPs are in standard form, that is, they must be minimized.} The results of \gencmsabip\ and \gencpcmsabip\ are presented in three columns for each algorithm. The first column (with heading 'Best') contains the best result obtained over 10 runs, the second column (with heading 'Avg.') shows the average of the best results obtained in each of the 10 runs, and the third column indicates the configuration (out of 4) that has produced the corresponding results. Finally, the results of \cplexopt\ and \cplexheur\ are both presented in two columns. The first column shows the value of the best feasible solution produced within the allowed computation time, and the second column shows the best gap (in percent) at the end of each run. Note that a gap of 0.0 indicates that optimality was proven. \\

\begin{sidewaystable}[p]
\caption{Numerical results for the 30 considered BIPs.}
\label{tab:results:bips}
\centering
\scalebox{0.78}{
\begin{tabular}{l@{\hskip -45pt}rrrrrrrrrrrrrrrr}
\hline
Instance && Opt.~Val. && \multicolumn{3}{c}{\textsc{\gencmsabip}} && \multicolumn{3}{c}{\textsc{\gencpcmsabip}} && \multicolumn{2}{c}{\textsc{Cplex-Opt}} && \multicolumn{2}{c}{\textsc{Cplex-Heur}} \\ \cline{5-7} \cline{9-11} \cline{13-14} \cline{16-17}
         &&          && \textbf{Best} & \textbf{Avg.} & \textbf{Conf.} && \textbf{Best} & \textbf{Avg.} & \textbf{Conf.} && \textbf{Value} & \textbf{Gap} && \textbf{Value} & \textbf{Gap} \\ \hline
\texttt{acc-tight5}          && 0.0            && 0.0           & \bf0.0           & 4     && 0.0           & \bf0.0           & 4 && \bf0.0           & 0.0    && \bf0.0           & 0.0 \\
\texttt{air04}               && 56137.0        && 56137.0       & \bf56137.0       & 4     && 56137.0       & \bf56137.0       & 4 && \bf56137.0       & 0.0    && \bf56137.0       & 0.0 \\
\texttt{cov1075}             && 20.0           && 20.0          & \bf20.0          & 3     && 20.0          & \bf20.0          & 4 && \bf20.0          & 0.0    && \bf20.0          & 0.0 \\
\texttt{eilB101}             && 1216.92        && 1216.92       & \bf1216.92       & 4     && 1216.92       & \bf1216.92       & 3 && \bf1216.92       & 0.0    && \bf1216.92       & 0.0 \\
\texttt{ex9}                 && 81.0           && -- --         & -- --            & -- -- && 81.0          & \bf81.0          & 2 && \bf81.0          & 0.0    && \bf81.0          & 0.0 \\
\texttt{netdiversion}        && 242.0          && 308.0         & 386.3            & 3     && 242.0         & 288.0            & 4 && 276.0            & 15.5   && \bf242.0         & 0.0 \\ 
\texttt{opm2-z7-s2}          && -10280.0       && -10280.0      & \bf-10280.0      & 4     && -10280.0      & \bf-10280.0      & 3 && \bf-10280.0      & 0.0    && \bf-10280.0      & 0.0 \\
\texttt{tanglegram1}         && 5182.0         && 5182.0        & \bf5182.0        & 4     && 5182.0        & \bf5182.0        & 4 && \bf5182.0        & 0.0    && \bf5182.0        & 0.0  \\
\texttt{vpphard}             && 5.0            && 5.0           & \bf5.0           & 3     && 5.0           & 5.5              & 3 && \bf5.0           & 0.0    && \bf5.0           & 100.0 \\ \hline
\texttt{ivu52}               && 481.007        && 657.6         & 884.23           & 3     && 16860.0       & 16860.0          & 3 && 647.05           & 25.77  && \bf490.09        & 2.0 \\
\texttt{opm2-z12-s14}        && -64291.0       && -63848.0      & \bf-62941.5      & 3     && -64149.0      & -61286.2         & 3 && -50200.0         & 78.62  && -44013.0      & 106.52 \\
\texttt{p6b}                 && -63.0          && -63.0         & -62.6            & 3     && -63.0         & -61.9            & 4 && -62.0            & 9.38   && \bf-63.0         & 8.08 \\
\texttt{protfold}            && -31.0          && -29.0         & \bf-27.4         & 3     && -30.0         & -27.0            & 3 && -25.0            & 43.53  && -26.0         & 54.82 \\
\texttt{queens-30}           && -40.0          && -39.0         & \bf-38.8         & 3     && -38.0         & -37.7            & 3 && -38.0            & 83.0   && -38.0         & 83.56 \\
\texttt{reblock354}          && -39280521.23   && -39261319.4   & -39246477.9      & 4     && -39271317.72  & -39262125.07     & 4 && \bf-39277743.28  & 0.27   && -39259518.82  & 0.53 \\
\texttt{rmine10}             && -1913.88       && -1911.39      & -1909.05         & 4     && -1911.75      & \bf-1910.72      & 3 && -1909.29         & 0.54   && -1910.49      & 0.75 \\
\texttt{seymour-disj-10}     && 287.0          && 287.0         & \bf287.5         & 3     && 287.0         & 287.6            & 4 && 288.0            & 2.08   && 288.0         & 2.17 \\
\texttt{wnq-n100-mw99-14}    && 259.0          && 259.0         & 268.2            & 3     && 268.0         & 272.6            & 4 && 275.0            & 13.98  && \bf267.0         & 11.92 \\ \hline
\texttt{bab1}                && ?              && -218764.89    & \bf-218764.89    & 3     && -218764.89    & \bf-218764.89    & 4 && \bf-218764.89    & 1.98   && \bf-218764.89    & 3.13 \\
\texttt{methanosarcina}      && ?              && 2730.0        & \bf2735.1        & 3     && 2841.0        & 2918.2           & 2 && 3080.0           & 56.79  && 2772.0        & 56.02 \\
\texttt{ramos3}              && ?              && 233.0         & 236.6            & 1     && 232.0         & \bf235.3         & 1 && 248.0            & 40.99  && 263.0         & 44.38 \\
\texttt{rmine14}             && ?              && -4266.31      & \bf-4254.25      & 3     && -4210.99      & -4182.77         & 3 && -962.12          & 347.84 && -1570.21      & 174.53 \\
\texttt{rmine25}             && ?              && -10297.66     & \bf-8836.25      & 3     && -1790.65      & -1130.70         & 4 && 0.0              & $\inf$ && 0.0           & $\inf$ \\
\texttt{sts405}              && ?              && 342.0         & \bf344.0         & 2     && 343.0         & 346.2            & 1 && 349.0            & 50.38  && 348.0         & 60.06 \\ 	
\texttt{sts729}              && ?              && 646.0         & \bf647.90        & 1     && 650.0         & 651.4            & 1 && 661.0            & 61.25  && 729.0         & 65.65 \\
\texttt{t1717}               && ?              && 178443.0      & \bf182489.3      & 2     && 188527.0      & 192958.1         & 1 && 200300.0         & 32.18  && 192942.0      & 29.92 \\
\texttt{t1722}               && ?              && 120946.0      & 124910.0         & 4     && 119478.0      & 126627.0         & 3 && 119086.0         & 14.81  && \bf118608.0      & 15.24 \\ \hline
\texttt{mcsp-2000-4}         && ?              && 514.0         & \bf519.5         & 2     && 527.0         & 527.0            & 3 && 527.0            & 100.0  && 527.0         & 100.0\\
\texttt{rflcs-2048-3n-div-8} && ?              && -115.0        & -105.2           & 1     && -114.0        & \bf-105.3        & 4 && 0.0              & inf    && 0.0           & inf \\ 
\texttt{rcjs-20testS6}       && ?              && -- --         & -- --            & -- -- && 11555.9       & \bf15279.3       & 1 && 20692.4          & 68.9   && -- --         & -- -- \\   \hline
\end{tabular}}
\end{sidewaystable}

\begin{figure}[!h]
\centering
\subfloat[Instance \texttt{air04}.\label{fig:comparison:air04}]{
  \includegraphics[width=0.48\textwidth]{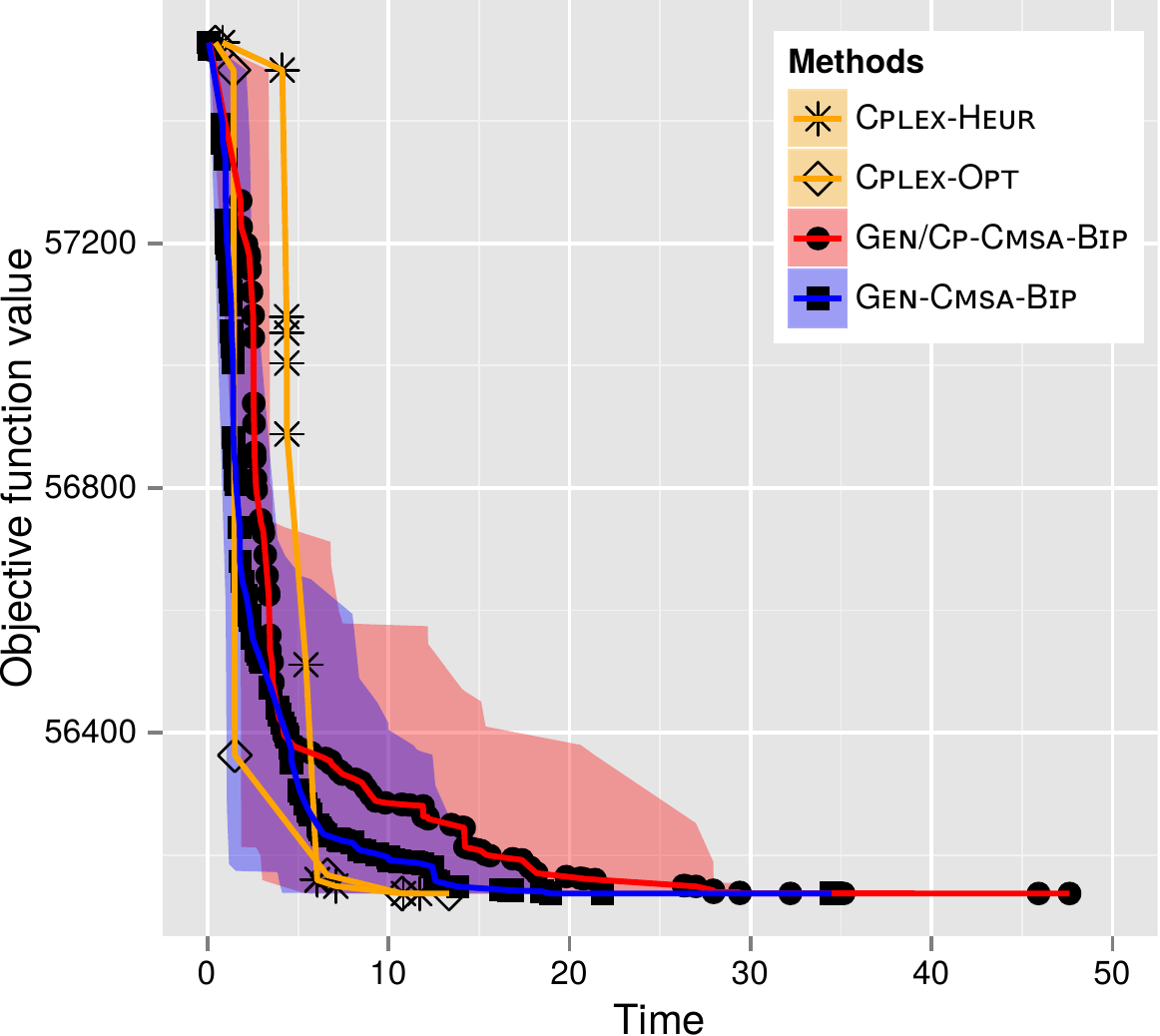}
} 
\subfloat[Instance \texttt{opm2-z12-s14}.\label{fig:comparison:opm2-z12-s14}]{
  \includegraphics[width=0.48\textwidth]{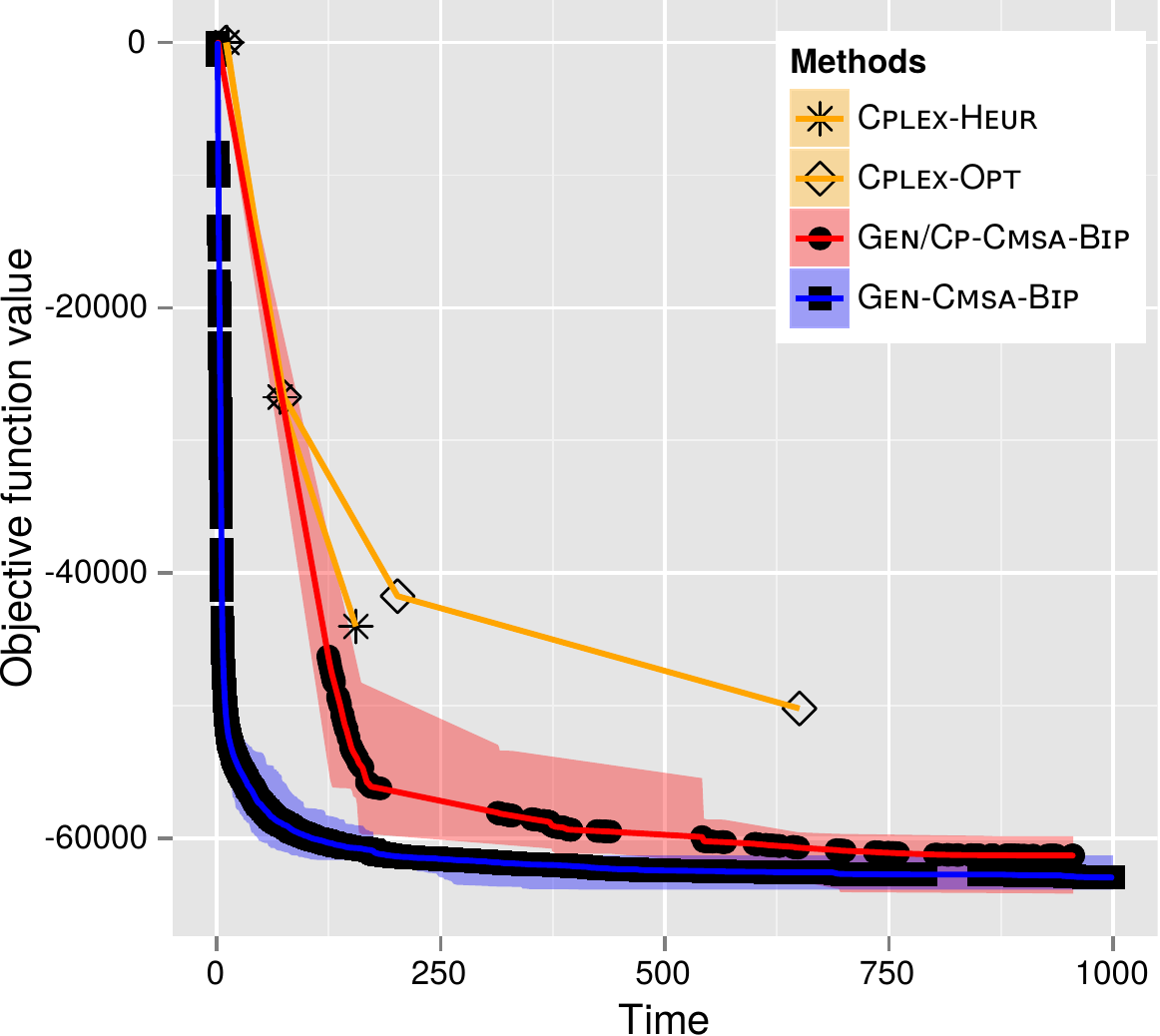}
} \\ 
\subfloat[Instance \texttt{protfold}.\label{fig:comparison:protfold}]{
  \includegraphics[width=0.48\textwidth]{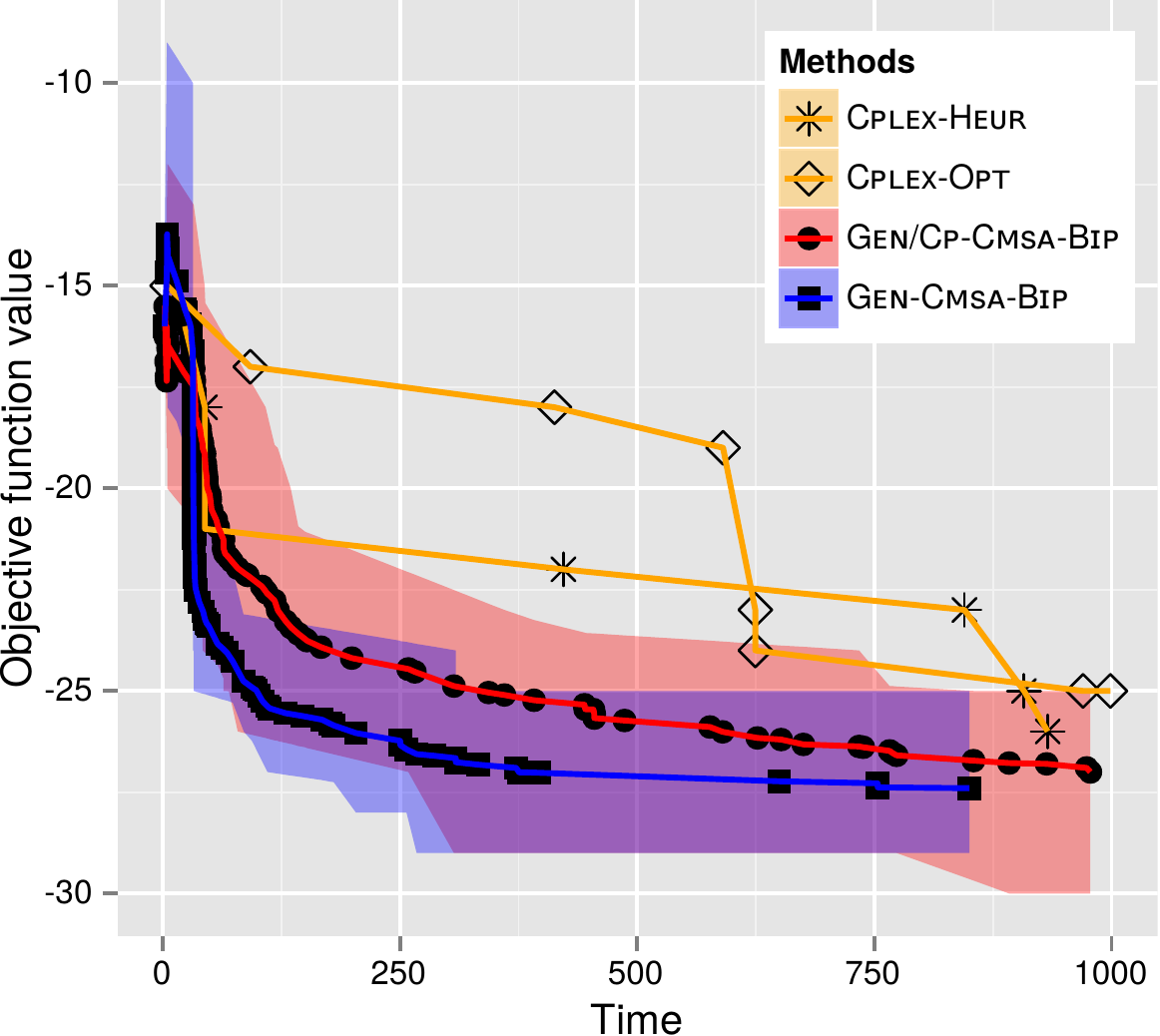}
} 
\subfloat[Instance \texttt{rmine14}.\label{fig:comparison:rmine14}]{
  \includegraphics[width=0.48\textwidth]{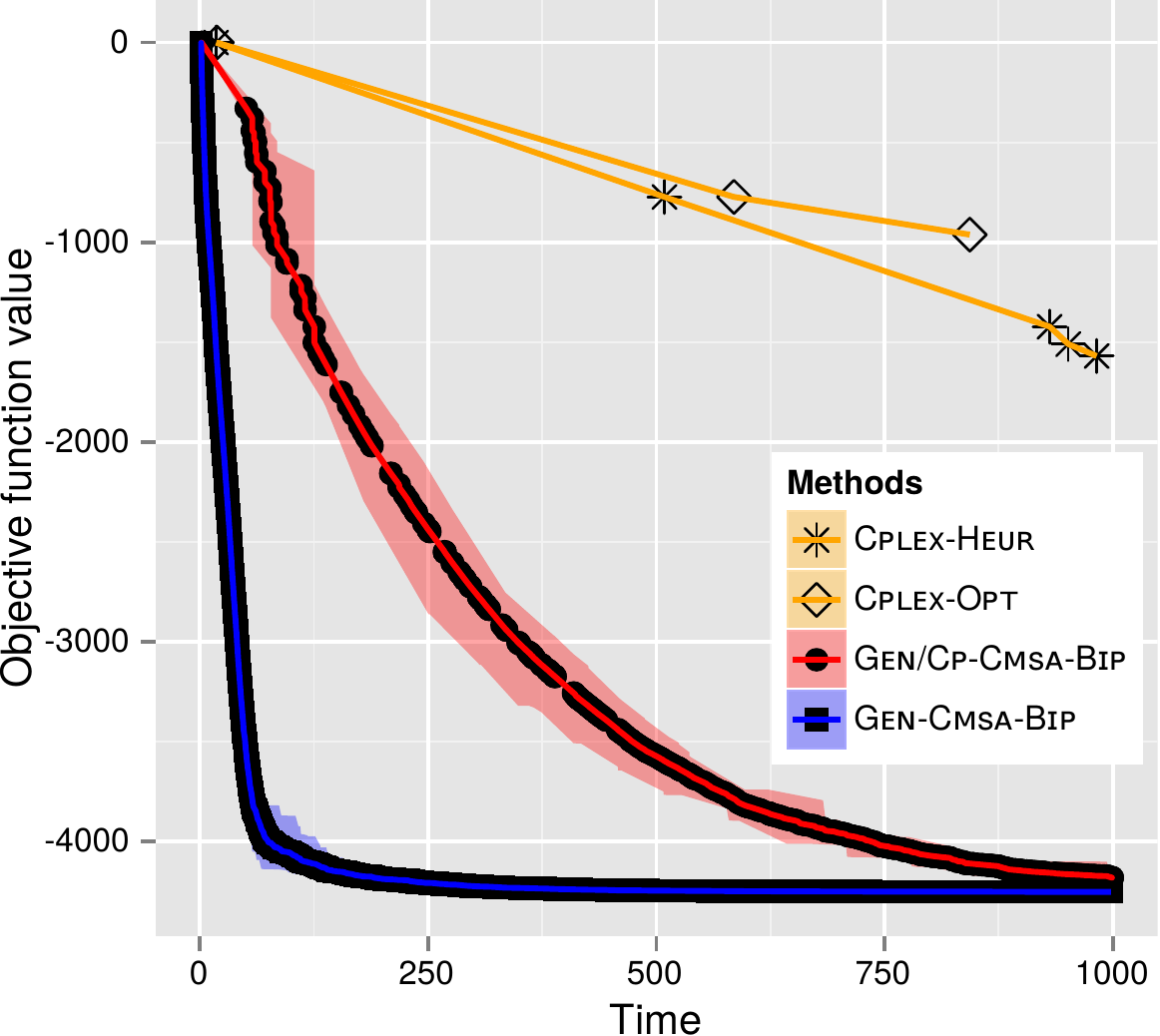}
} \\ 
\subfloat[Instance \texttt{t1717}.\label{fig:comparison:t1717}]{
  \includegraphics[width=0.48\textwidth]{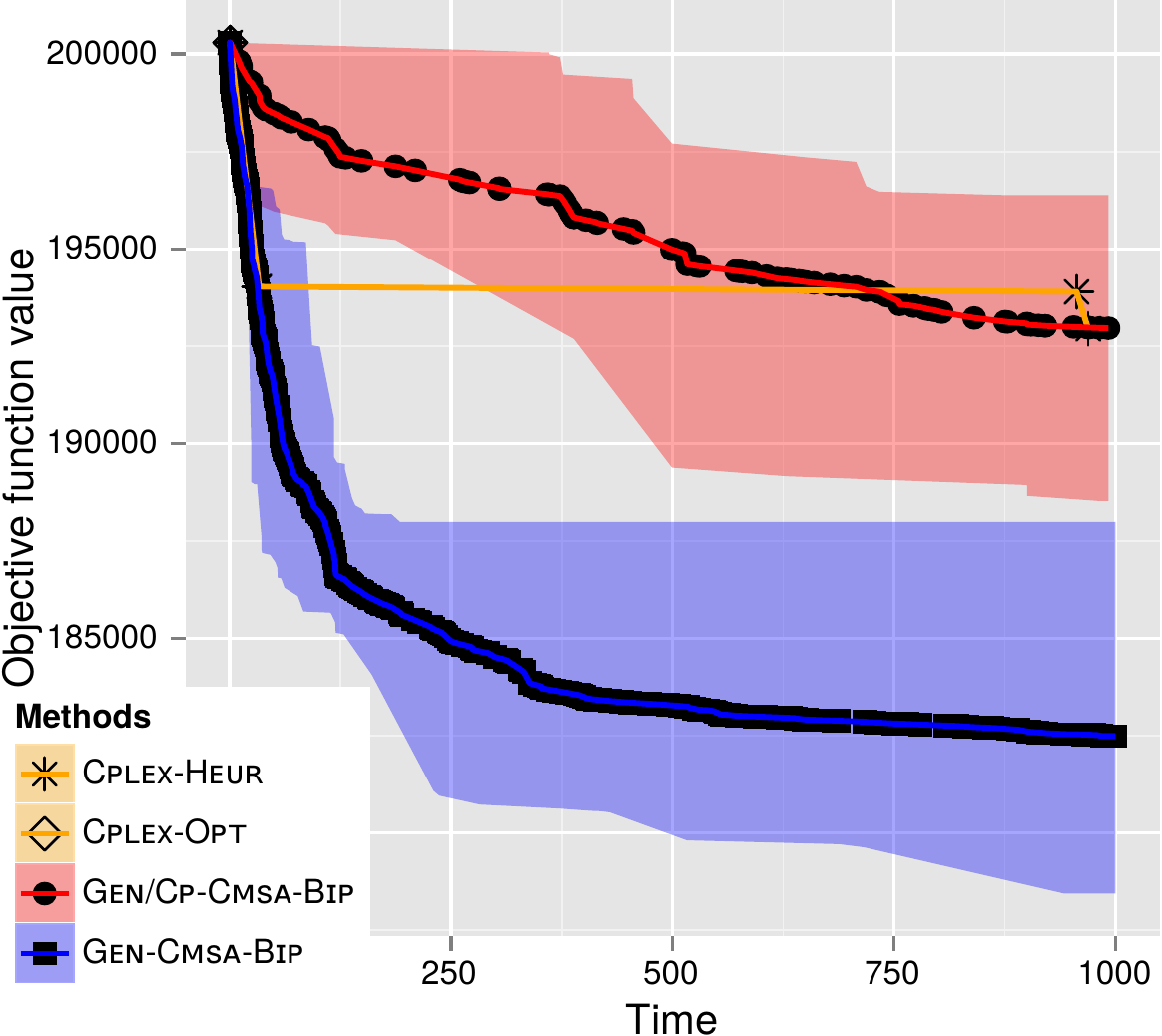}
} 
\subfloat[Instance \texttt{rflcs-2048-3n-div-8}.\label{fig:comparison:rflcs}]{
  \includegraphics[width=0.48\textwidth]{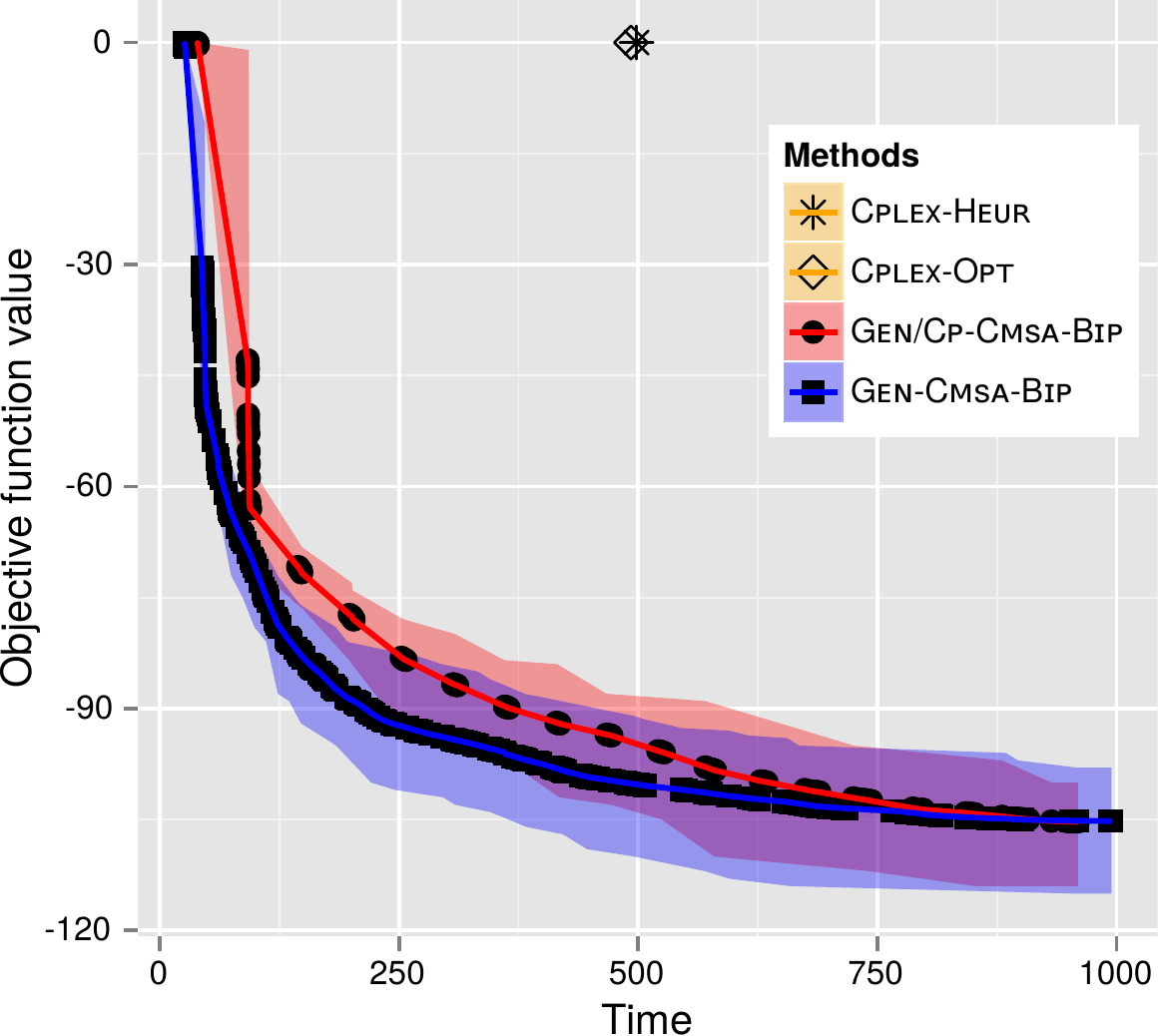}
}
\caption{Anytime performance of \gencmsabip\ and \gencpcmsabip\ in comparison to the two CPLEX variants. The performance of the two generic CMSA versions is shown via the mean performance together with the confidence ribbon (based on 10 independent runs).}
\label{fig:comparison}
\end{figure}

\noindent The following observations can be made:
\begin{itemize}
  \item Concerning the BIPs classified as \emph{easy} (see the first nine table rows), it can be noticed that \cplexheur\ always generates an optimal solution, even though optimality can not be proven in two cases. \gencpcmsabip--that is, the generic CP-supported CMSA variant---also produces an optimal solution in at least one out of 10 runs for all nine problem instances. However, in three cases, the algorithm fails to produce an optimal solution in all 10 runs per instance. The results of the basic generic CMSA variant (\gencmsabip) are quite similar. However, for instance \texttt{ex9} it is not able to produce any feasible solution, and for instance \texttt{netdiversion} the results are clearly inferior to those of \gencpcmsabip. Nevertheless, the support of CP also comes with a cost. This can be seen when looking at the anytime behaviour of the algorithms as shown for six exemplary cases in Figure~\ref{fig:comparison}. In particular, \gencpcmsabip\ is often not converging as fast to good solutions as \gencmsabip. 
  \item The increased difficulty of the instances labelled as \emph{hard} (see table rows 10-18), produces more differences between the four approaches. In fact, sometimes one of the CPLEX variants is clearly better than the two CMSA versions (see, for example, instance \texttt{ivu52}), and sometimes the generic CMSA variants outperform the CPLEX versions (such as, for example, for instance \texttt{opm2-z12-s14}). As the CP-support is more costly for these instances, the results of \gencmsabip\ are generally a bit better than those of \gencpcmsabip. The effect of the increased cost of the CP support can also nicely be observed in the anytime behaviour of the algorithms for two hard instances in Figure~\ref{fig:comparison:protfold} and Figure~\ref{fig:comparison:opm2-z12-s14}.
  \item In the context of the nine \emph{open} instances, the generic CMSA variants clearly outperform the standalone application of CPLEX, with the exception of instance \texttt{t1722}. The same holds for the three additional, difficult problem instances (last three table rows). Note, especially, that for instance \texttt{rcjs-20testS6} the support of CP pays off again, as it is difficult to find feasible solutions for this instance.  
\end{itemize}

\begin{figure}[!h]
\centering
\subfloat[Instance \texttt{eilB101}.\label{fig:configuration-performance:eilB101}]{
  \includegraphics[width=0.7\textwidth]{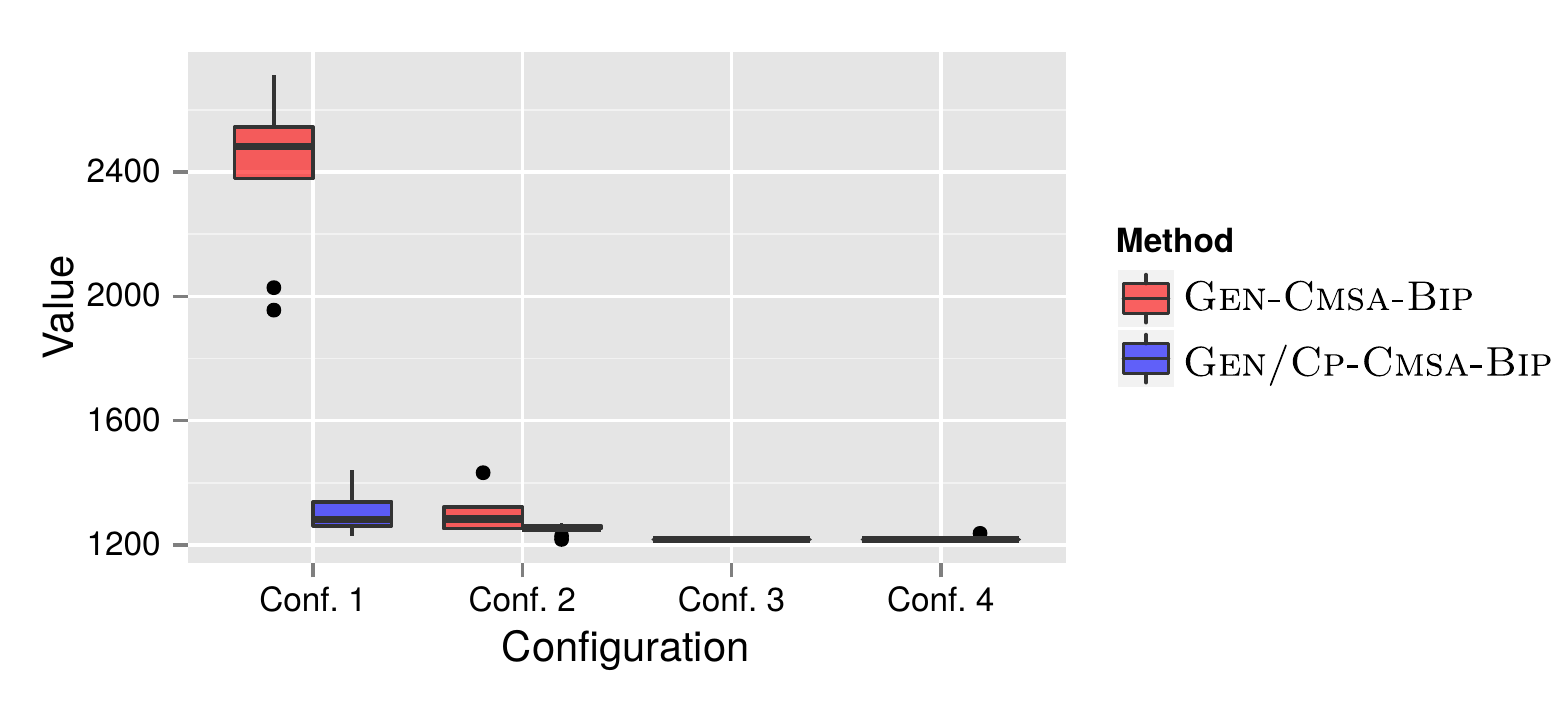}
} \\
\subfloat[Instance \texttt{opm2-z7-s2}.\label{fig:configuration-performance:opm2-z7-s2}]{
  \includegraphics[width=0.7\textwidth]{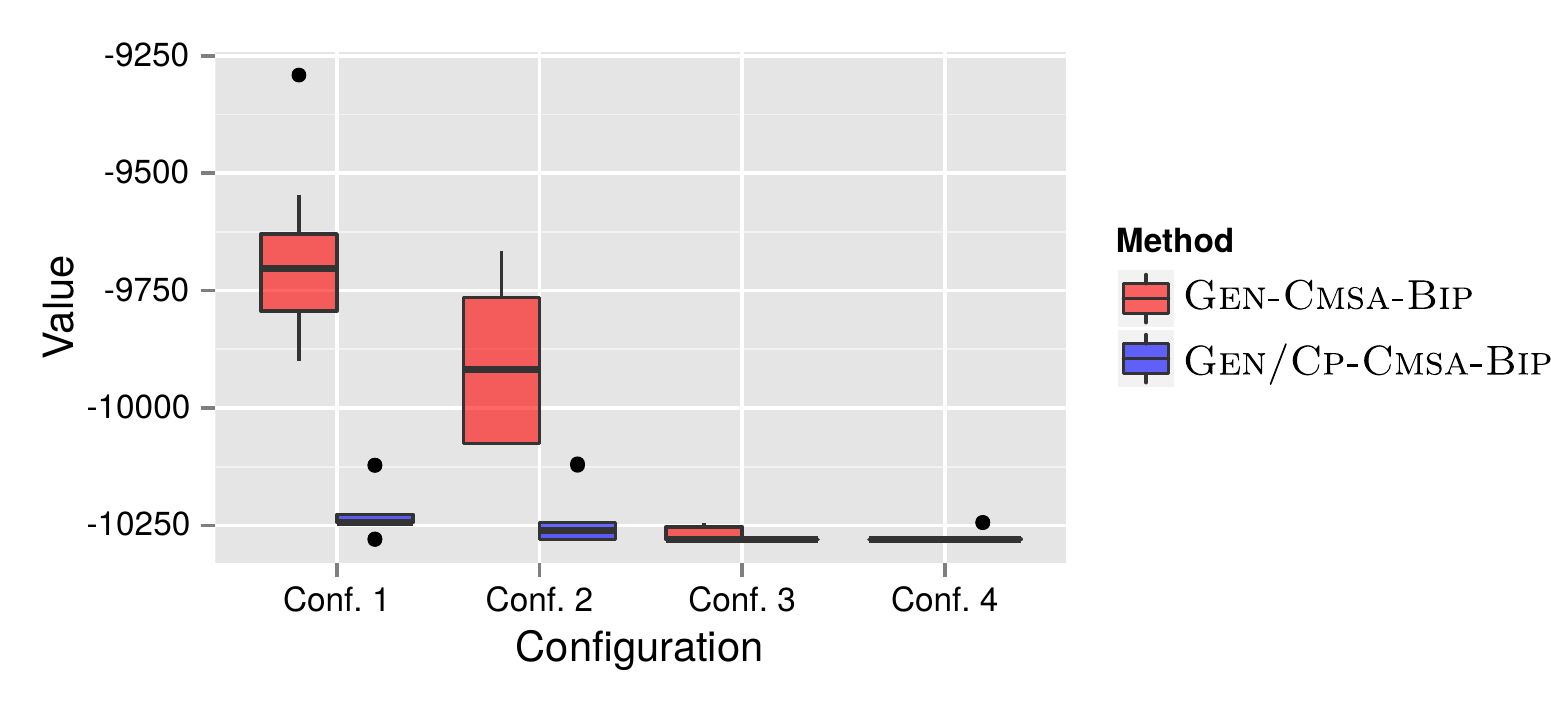}
} \\
\subfloat[Instance \texttt{rflcs-2048-3n-div-8}.\label{fig:configuration-performancen:2048_3n-div-8.0}]{
  \includegraphics[width=0.7\textwidth]{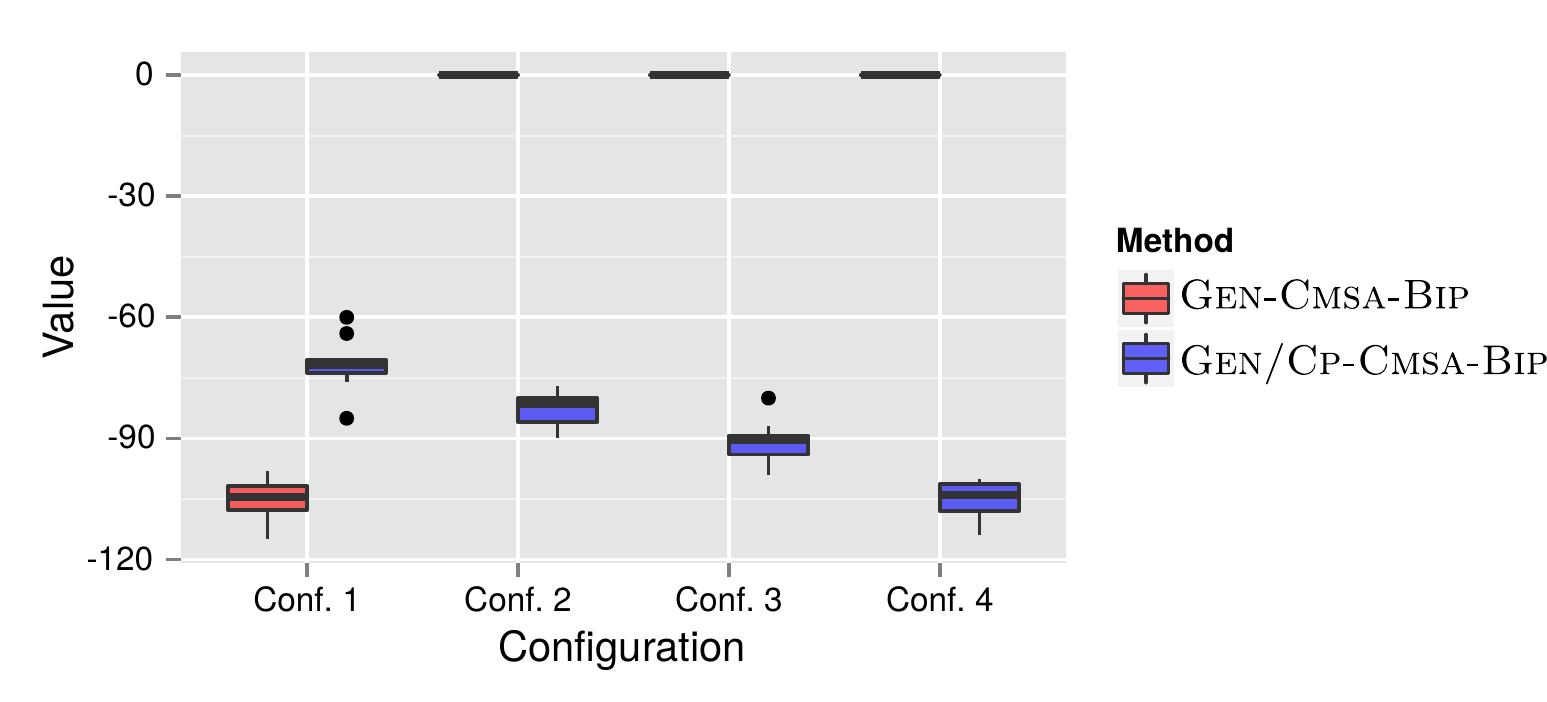}
} 
\caption{Boxplots showing the 10 results per algorithm configuration for \gencmsabip\ and \gencpcmsabip\ in the context of three of the considered problem instances.}
\label{fig:configuration-performance}
\end{figure}

Summarizing, with increasing problem size/difficulty, the advantage of the generic CMSA variants over the standalone application of CPLEX becomes more and more pronounced. However, as the cost of the CP support also increases with growing problem size, \gencpcmsabip\ is only able to outperform \gencmsabip\ when finding feasible solutions is really difficult. However, we noticed that the CP support has also an additional effect, which is shown in the graphics of Figure~\ref{fig:configuration-performance}. Each boxplot shows the final results (obtained by 10 runs per instance) of each of the four parameter configurations for both generic CMSA variants. Interestingly, the use of CP during solution construction flattens out the quality differences between the four parameter configurations. This can be seen in all three boxplots. In Figure~\ref{fig:configuration-performance:opm2-z7-s2}, for example, the results of \gencmsabip\ are good with configurations 3 and 4, while they are significantly worse with configurations 1 and 2. In contrast, the results of \gencpcmsabip, while also being best with configurations 3 and 4, are only slightly worse with configurations 1 and 2. In that sense, the CP-support makes the algorithm more robust with respect to the parameter setting. 

\section{Conclusions}
\label{sec:conclusions}

In this work, we developed a problem-agnostic CMSA algorithm for solving binary linear integer programs. The main challenge was on constructing solutions to unknown problems in such a way that feasibility is quickly reached. For this purpose, in addition to the basic algorithm, we developed an algorithm variant that makes use of a constraint programming tool. Concerning the results, we were able to observe that the use of the constraint programming tool pays off for those instances for which it reaching feasiblity is rather difficult. In general, with growing problem size and/or difficulty, both CMSA variants have an increasing advantage over the standalone appliaction of the ILP solver CPLEX. In a sense, our generic CMSA can be seen as \emph{a better way of using an ILP solver} in many cases. Concerning future work, we plan to extend our work towards general ILPs. Moreover, we plan to work on a mechanism for automatically adjusting the algorithm parameters at the start of a run.



\end{document}